\documentclass[10pt,twocolumn,letterpaper]{article}

\usepackage{3dv}
\usepackage{times}
\usepackage{epsfig}
\usepackage{graphicx}
\usepackage{amsmath}
\usepackage{amssymb}
\usepackage{fancychapters}
\usepackage[autostyle]{csquotes}  

\usepackage[pagebackref=true,breaklinks=true,letterpaper=true,colorlinks,bookmarks=false]{hyperref}

\usepackage{siunitx}
\usepackage{subfigure}
\usepackage{diagbox}
\usepackage{multirow}

\def\mean{\mathop{\operator@font mean}}

\usepackage[nocompress]{cite}

\def\ie{\emph{i.e.}}
\def\eg{\emph{e.g.}}

\def\etal{\emph{et al.}}

\threedvfinalcopy 

\ifthreedvfinal\fi
\begin{document}

\title{Learning to Generate Facial Depth Maps}

\author{Stefano Pini \quad Filippo Grazioli \quad Guido Borghi \quad Roberto Vezzani \quad Rita Cucchiara\\
Department of Engineering ``Enzo Ferrari''\\
University of Modena and Reggio Emilia\\
via Vivarelli 10, Modena 41125, Italy\\
\{name.surname\}@unimore.it}

\maketitle

\begin{abstract}
In this paper, an adversarial architecture for facial depth map estimation from monocular intensity images is presented. 
By following an image-to-image approach, we combine the advantages of supervised learning and adversarial training, proposing a conditional Generative Adversarial Network that effectively learns to translate intensity face images into the corresponding depth maps.
Two public datasets, namely Biwi database and Pandora dataset, are exploited to demonstrate that the proposed model generates high-quality synthetic depth images, both in terms of visual appearance and informative content. 
Furthermore, we show that the model is capable of predicting distinctive facial details by testing the generated depth maps through a deep model trained on authentic depth maps for the face verification task.
\end{abstract}


\section{Introduction}
Depth estimation is a task at which humans naturally excel thanks to the presence of two high-quality stereo cameras (\ie~the human eyes) and an exceptional learning tool (\ie~the human brain). 
What makes humans so excellent at estimating depth even from a single monocular image and how does this learning process happen?
One hypothesis is that we develop the faculty to estimate the 3D structure of the world through our past visual experience, which consists in an extremely large number of observations associated with tactile stimuli (for small objects) and movements (for wider spaces)~\cite{zhou2017unsupervised}. This process allows humans to develop the capability to infer the structural model of objects and scenes they see, even from monocular images. \\
Even though depth estimation is a natural human brain activity, the task is an ill-posed problem in the computer vision context, since the same 2D image may be generated by different 3D maps. Moreover, the translation between these two domains is demanding due to the extremely different source of information that belong to intensity images and depth maps: texture and shape data, respectively.
\\
Traditionally, the computer vision community has broadly addressed the problem of depth estimation in different ways, as \textit{Stereo Cameras}~\cite{konda2013unsupervised, yamaguchi2012continuous}, \textit{Structure from Motion} \cite{ding2017fusing,cavestany2015improved}, and \textit{Depth from shading and light diffusion}~\cite{woodham1980photometric,tao2015depth}.
\\
The mentioned methods suffer from different issues, like depth homogeneity and missing values (resulting in holes in depth images). Additional challenging elements are related to the camera calibration, setup, and post-processing steps that can be time consuming and computational expensive.
Recently, thanks to the advances of deep neural networks, the research community has investigated the monocular depth estimation task from intensity images in order to overcome to previously reported issues~\cite{atapour2018real, godard2017unsupervised, zhou2017unsupervised, garg2016unsupervised, eigen2014depth}.

\begin{figure}[t!]
  \centering
  \includegraphics[width=0.9\columnwidth]{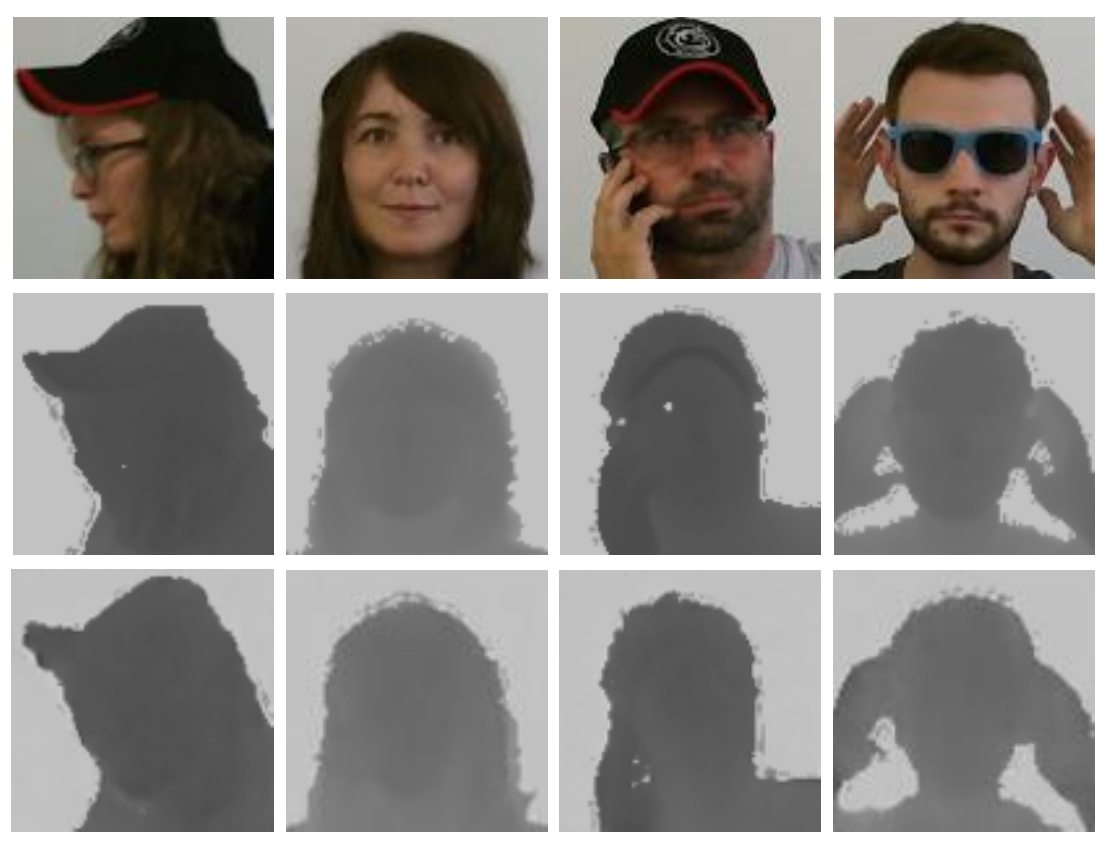} 
  \caption{Examples of the RGB face images (first row), ground-truth depth images (second row), and in the last row the depth maps estimated by the proposed model. Not only are generated face images visually realistic and pleasant, but they also preserve discriminative features for the face verification task.}
  \label{fig:intro}
\end{figure}

\begin{figure*}[t!]
  \centering
  \includegraphics[width=1\linewidth]{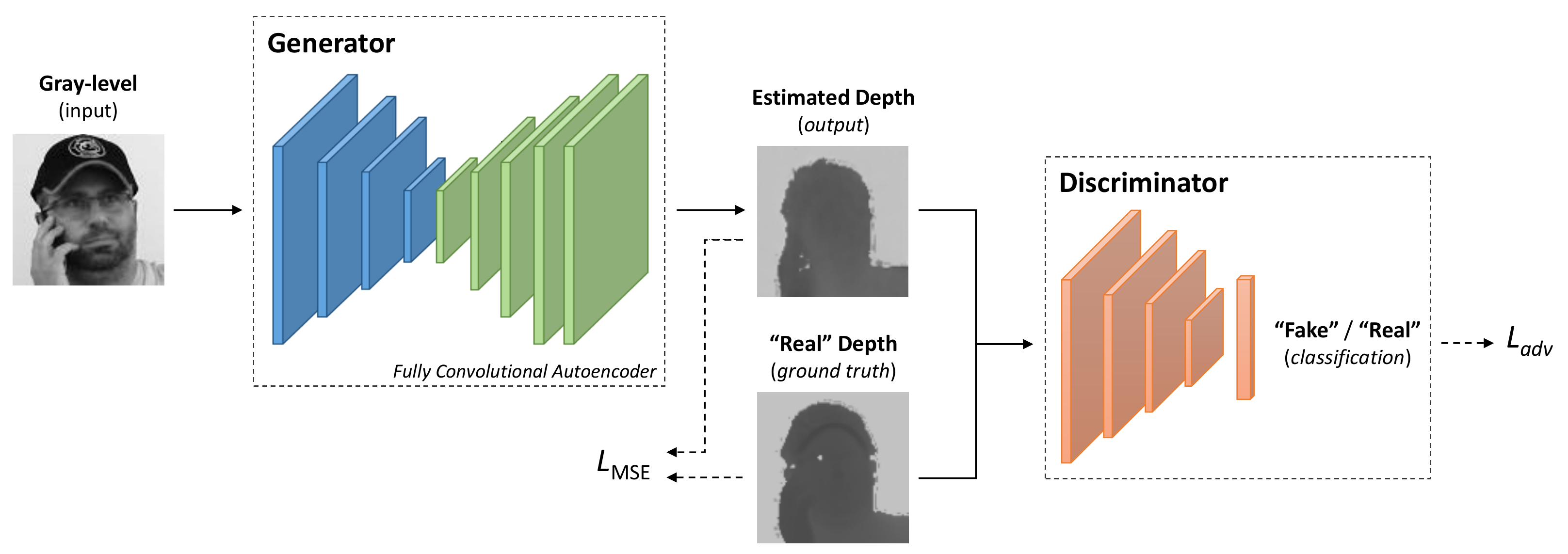} 
  \caption{Overall view of the proposed Conditional GAN architecture. The architecture of the \textit{Generator} is: k5n128s2 - k5n256s2 - k5n512s2 - k5n1024s2 - k5n512s2 - k5n256s2 - k5n128s2 - k5n64s2 - k5n1s1. The architecture of the \textit{Discriminator} is: k5n128s2 - k5n256s2 - k5n512s2 - k5n1024s2 - fc1. 
$k$, $n$, $s$, $fc$ correspond to the kernel size, the number of feature maps, the stride, and the number of fully connected units, respectively.} 
  \label{fig:architecture}
\end{figure*}

This paper presents a framework for the generation of depth maps from monocular intensity images of human faces. 
An adversarial approach~\cite{goodfellow2014generative,mirza2014conditional} is employed to effectively train a fully convolutional autoencoder that is able to estimate facial depth maps from the corresponding gray-level images.
To train and test the proposed method, two public dataset, namely \textit{Pandora}~\cite{borghi17cvpr} and \textit{Biwi Kinect Head Pose} \cite{fanelli_IJCV} dataset, that consists of a great amount of paired depth and intensity images, are exploited. 
To the best of our knowledge, this is one of the first attempts to tackle this task through an adversarial approach that, differently from global depth scene estimation, involves small sized objects, full of details: the human faces.
\\
Finally, we investigate how to effectively measure the performance of the system, introducing a variety of \textit{pixel-wise} metrics. 
Besides, we introduce a \textit{Face Verification} model trained on the original face depth images, to check if the generated images maintain the facial distinctive features of the original subjects, not only when visually inspected by humans, but also when processed by deep convolutional networks.

\section{Related Works}
We consider related work within two distinct domains: \textit{Monocular Depth Estimation} and \textit{Domain Translation}, detailed in the following sections.\\

\noindent \textbf{Monocular Depth Estimation.}
Facial depth estimation from monocular images has been investigated during the last decade.
In \cite{sun2011depth}, \textit{Constrained Independent Component Analysis} is exploited for depth estimation from various pose of 2D faces. A nonlinear least-square model is employed in~\cite{sun2013depth} to predict the 3D structure of the human face. Both methods rely on an initial face parameters detection based on facial landmarks. Consequently, the final estimation is influenced by detection performance and, therefore, head pose angles. 
In \cite{zheng2008robust}, this task is tackled as a statistical learning problem, using the \textit{Local Binary Patterns} as feature, but only frontal faces are taken into account. Reiter~\etal~\cite{reiter2006estimation} propose the use of canonical correlation analysis to predict depth information from RGB frontal face images. In~\cite{kong2016effective}, the face \textit{Delaunay Triangulation} is exploited in order to estimate depth maps based on similarity. 
Cui \etal \cite{cui2018improving} propose a cascade FCN and a CNN architecture to estimate the depth information. In particular, the FCN aims to recover the depth from an RGB image, while the CNN is employed during the training phase to maintain the original subject's identity. 
\\
A wide body of literature addresses the monocular depth estimation task in the automotive context~\cite{godard2017unsupervised,atapour2018real} or in indoor scenarios~\cite{li2017two,xu2017multi}.
The \textit{Markov Random Field} (MRF) and the linear regression are employed in~\cite{saxena2006learning} to predict depth values. An evolution of the MRF combines the 3D orientation and position of segmented patches within RGB images~\cite{saxena2009make3d}. The main challenges to these works is that depth values are locally predicted therefore scene depth prediction lacks of global coherence.
To improve the global scene depth prediction accuracy, sparse coding has been investigated in \cite{baig2014im2depth}, while semantic labels are exploited in \cite{ladicky2014pulling}.
\\
Recently, this research field has received a great improvement thanks to the introduction of Convolutional Neural Networks~\cite{eigen2014depth,eigen2015predicting,liu2016learning}.
Several works propose the use of RGB samples paired with depth images as ground truth data in order to learn how to estimate depth maps by means of a supervised approach. In \cite{eigen2014depth,eigen2015predicting}, a two-scale network is trained on intensity images to produce depth values. The main issue is related to the limited size of publicly-available training data and the overall low image quality~\cite{li2015depth,laina2016deeper}.
\\
In this work, we aim to investigate the adversarial training approach~\cite{goodfellow2014generative} in order to propose a method that directly estimate facial depth maps, without a-priori facial feature detection, like facial landmarks or head pose angles.\\

\noindent \textbf{Domain Translation.}
The \textit{domain translation} task, which is often referred as \textit{image translation} in the computer vision community, consists in learning a parametric mapping function between two distinct domains. 
\\
Image-to-image translation problems are often formulated as a per-pixel classification or regression \cite{long2015fully, xie2015holistically, iizuka2016let, larsson2016learning, zhao2016energy}. Borghi \etal~\cite{borghi17cvpr} propose an approach for computing the appearance of a face using the corresponding depth information based on a traditional CNN combining aspects of autoencoders \cite{masci2011stacked} and Fully Convolutional Networks \cite{long2015fully}.
Recently, a consistent body of literature has addressed the image-to-image translation problem by exploiting \textit{conditional Generative Adversarial Networks} (cGANs) \cite{mirza2014conditional} in order to learn a mapping between two image domains. 
Wang \etal~\cite{wang2016generative} proposed a method, namely \textit{Style GAN}, that renders a realistic image from a synthetic one.
Isola \etal~\cite{isola2017image} demonstrated that their model, called \textit{pix2pix}, is effective at synthesizing photos from semantic labels, reconstructing objects from edge maps, and colorizing images. 
In \cite{liu2016coupled}, a framework of coupled GANs, which is able to generate pairs of corresponding images in two different domains, was proposed.\\
We tackle the domain translation task in order to generate facial depth maps which are visually pleasant and contain enough discriminative information for the face verification task.

\section{Proposed Method} \label{sec:proposedmethod}
In this section, we present the proposed model for depth estimation from face intensity images, detailing the cGAN architecture (Section \ref{subsec:gan-training}), its training procedure (Section \ref{sec:adversarial-training}), and the adopted pre-processing face crop algorithm (Section \ref{sec:face-crop}).
The implementation of the model follows the guidelines proposed in \cite{goodfellow2014generative}.

\subsection{Depth Estimation Model}
Following the work of Goodfellow \etal~\cite{goodfellow2014generative} and Mirza \etal~\cite{mirza2014conditional}, the proposed architecture is composed of a generative network $G$ and a discriminative network $D$. $G$ corresponds to an estimation function that predicts the depth map $I^{gen} = G(I^{gray})$ of a given face gray-level image $I^{gray}$, while $D$ corresponds to a discriminative function that distinguishes between original (\ie~\enquote{real}) and generated (\ie~\enquote{fake}) depth maps.

\subsubsection{Network architecture} \label{subsec:gan-training}
\noindent \textbf{Generator.}
The generator network is based on the fully convolutional architecture depicted in Figure \ref{fig:architecture} that, following the paradigm of \textit{conditional} GANs, takes a face intensity image as input and estimates the corresponding depth map.
The first part of the network acts as an encoder, mapping the input image into a $1024$-dimensional embedding with a spatial size $16$ times smaller than the input one. It is composed of four convolutional layers with kernel size $5$, stride $2$, and $128$, $256$, $512$, and $1024$ features maps, respectively. Each layer is followed by the \textit{Leaky ReLU} activation function~\cite{maas2013rectifier} with a negative slope of $0.2$.\\
The second part of the network, acting as a decoder, generates a depth image by processing the face embedding produced by the encoder. It is composed of four transposed convolution layers (also known in the literature as fractionally-strided convolution layers) which increase the embedding resolution up to the original image size. 
The layers are applied with kernel size $5$, stride $2$, and $512$, $256$, $128$, and $64$ features maps, respectively, and they are followed by the \textit{ReLU} activation function. 
Then, a standard convolutional layer, followed by a hyperbolic tangent activation function (\textit{tanh}), produces the final depth map estimation.
Batch normalization is employed before each activation function, except the last one, for regularization purposes.\\

\noindent \textbf{Discriminator.}
The discriminator network, depicted in Figure~\ref{fig:architecture}, takes as input a depth map and predicts the probability of the input to be a real or a generated depth map.
The first part of the discriminator shares the same architecture with the encoder part of the generator network.
Then, the $1024$-dimensional embedding is flattened 
and a fully connected layer with one unit and a sigmoid activation function are applied obtaining a final score in the range $[0, 1]$ where $0$ corresponds to a \enquote{real} depth map and $1$ to a \enquote{fake} one.

\subsubsection{Adversarial training} \label{sec:adversarial-training}
During the training procedure, the discriminator network $D$, with parameters $\theta_{d}$, is trained to predict whether a depth map is \enquote{real} or \enquote{fake} by maximizing the probability of assigning the correct label to each sample.
Meanwhile, the generator network $G$, with parameters $\theta_{g}$, is trained in order to generate realistic depth maps and fool the discriminator $D$.
From a mathematical perspective, the training can be formalized as the optimization of the following min-max problem:
\begin{align}
	\label{eq:minmax}
\begin{aligned}
 	\min_{\theta_{g}} \max_{\theta_{d}}\ &\mathbb{E}_{x \sim p_{dpt}(x)}[\log (D(x))] \\
 	&+ \mathbb{E}_{y \sim p_{gray}(y)}[\log (1 - D(G(y)))]
\end{aligned}
\end{align}
\noindent where $D(x)$ is the probability of being a \enquote{real} depth image (consequently $1 - D(G(y))$ is the probability to be a \enquote{fake} depth image), $p_{dpt}$ is the distribution of the real depth maps, and $p_{gray}$ is the distribution of the intensity images.
This approach leads to a generative model which is capable of generating \enquote{fake} images that are highly similar to the \enquote{real} ones, thus indistinguishable by the discriminator $D$.

To reach this goal, the following loss functions are employed during the training with the \textit{Adam} optimizer~\cite{kingma2014adam} with initial learning rate of $20^{-4}$ and betas $0.5$ and $0.999$.
Regarding the discriminator network, a binary categorical cross entropy loss function, defined as
\begin{equation}
	\label{eq:adversarial_loss}
	L_{adv}(\mathbf{y}, \mathbf{t}) = - \frac{1}{N} \sum_{i=1}^N \left[ t_i \log y_i + (1 - t_i) \log (1 - y_i) \right]
\end{equation}
where $y_i = D(I_i)$ is the discriminator prediction regarding the $i$-th input depth map and $t_i$ is the corresponding ground truth, is applied to the discriminator output.\\
Regarding the generator network, we aim to generate images that are similar to the ground truth depth maps as well as capable of fooling the discriminator network (\ie~visually indistinguishable from real depth maps).
For fulfilling the first goal, we apply the Mean Squared Error (MSE) loss function:
\begin{equation}
	\label{eq:mse_loss}
	L_{MSE}(\mathbf{s}^{g}, \mathbf{s}^{d}) = \frac{1}{N} \sum_{i=1}^N{ \lVert G(s^{g}_i) - s^{d}_i \rVert^2_2}
\end{equation}
where $\mathbf{s}^{g}$ and $\mathbf{s}^{d}$ are respectively the input gray-level images and the target depth maps.
To accomplish the second goal, we feed the discriminator with generated depth images and apply the adversarial loss on the discriminator prediction to evaluate if the generated images are capable of fooling the discriminator network.\\ 
Then, we back-propagate the gradients up to the input of the generator network and update the generator parameters while keeping fixed the discriminator weights.
Therefore, we aim to solve the back-propagation problem minimizing: 
\begin{equation}
\label{eq:a}
	\hat{\theta}_g = \arg \min_{{\theta}_g} L_{G} \left( \mathbf{s}^{g}, \mathbf{s}^{d} \right)
\end{equation}
where $L_G$ is a combination of two components, defined as the following weighted sum:
\begin{equation}
	\label{eq:generator_loss}
	L_{G}(\mathbf{s}^{g}, \mathbf{s}^{d}) = \lambda \cdot L_{MSE}(\mathbf{s}^{g}, \mathbf{s}^{d}) + L_{adv}(G(\mathbf{s}^{g}), \mathbf{1})
\end{equation}
in which $\lambda$ is a weighting parameter that controls the impact of the $L_{MSE}$ loss with respect to the adversarial loss.

\subsection{Dynamic face crop} \label{sec:face-crop}
The head detection task is out of the scope of this paper, therefore a trivial dynamic face crop algorithm is adopted in order to accurately extract face bounding boxes from the considered datasets including a small portion of background.
In particular, given the head center position $(x_H, y_H)$ in a depth map (we assume that the head center position is provided in the dataset annotations), a bounding box of width $w_H$ and height $h_H$ is extracted defining its width and height as
\begin{equation}
	\label{eq:headBB}
	w_H=\frac{f_x \cdot R_x}{D} \ \ \quad h_H=\frac{f_y \cdot R_y}{D}
\end{equation}
where $R_x, R_y$ are the average width and height of a face (we consider $R_x=R_y=320$), $f_x, f_y$ are the horizontal and the vertical focal lengths in pixels of the acquisition device (in the considered case: $f_x=f_y=365$ and $f_x=f_y=370$ for \textit{Pandora} and \textit{Biwi} datasets, respectively), and $D$ is the distance between the head center and the acquisition device which is estimated averaging the depth values around the head center. 
Computed values are used to crop the face in both depth maps and intensity images.

\section{Experimental Results}
In this section, experimental results, obtained through a \textit{cross-subject evaluation}, are reported.
In particular, we investigate the use of pixel-wise metrics in order to verify the generation capability of the proposed adversarial model (Section \ref{sec:pixel wise metrics}).
Furthermore, we evaluate the quality of the estimated facial depth maps by means of a \textit{Face Verification} task (Section \ref{sec:face-verification}).
The code and the network models are publicly released\footnote{Link omitted for double-blind review.}.

\begin{figure}[t!]
  \centering
  \includegraphics[width=0.98\linewidth]{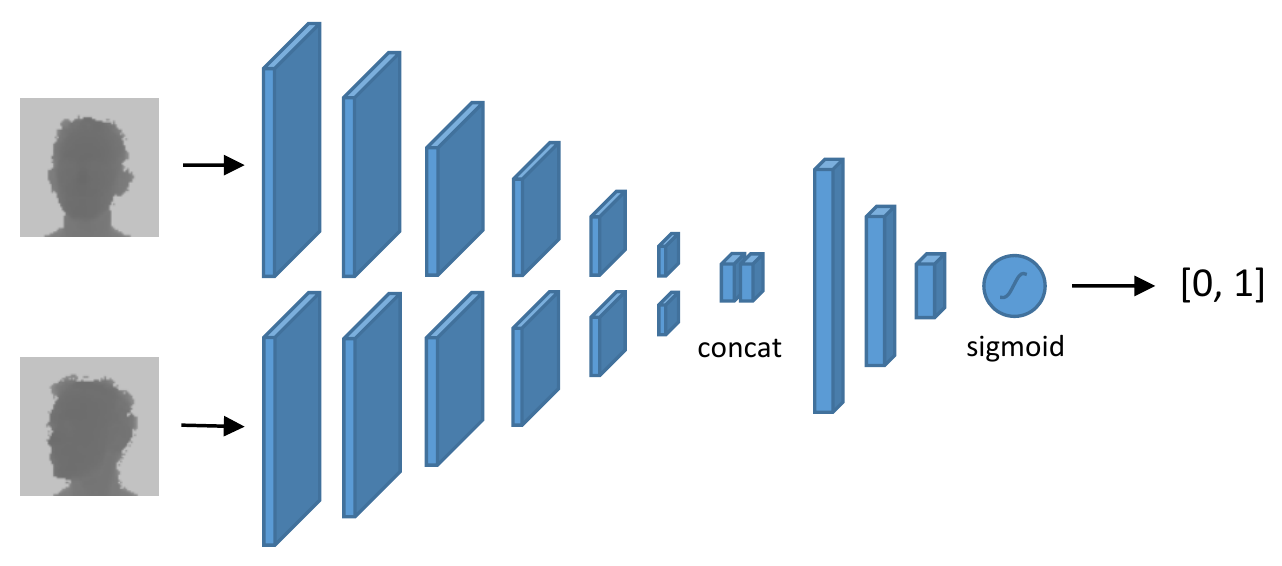} 
  \caption{Architecture of the Siamese network: k3n64s2 - k3n128s2 - k3n256s2 - k3n256s2 - k3n256s2 - avgpoolk2 - fc128 - fc32 - fc1. $k$, $n$, $s$, $fc$ correspond to kernel size, number of feature maps, stride, and fully-connected units. The output of the network is a continuous similarity score.}
  \label{fig:architecture_siam}
\end{figure}

{
\renewcommand{\arraystretch}{1.0}
\begin{table*}
      \begin{center}
          \begin{tabular}[t]{cl|cccc||cccc}
          	   & & \multicolumn{4}{c||}{\textbf{Pandora} \cite{borghi17cvpr}} & \multicolumn{4}{c}{\textbf{Biwi} \cite{fanelli_IJCV}} \\ 
              \textbf{Metrics} & & \textit{cGAN} 		& \textit{AE}	& \cite{isola2017image}	 & \cite{cui2018improving}			& \textit{cGAN}		& \textit{AE}	& \cite{isola2017image}  & \cite{cui2018improving}\\  \hline
              \rule{0pt}{2ex}
              $L_1$ Norm						&\multirow{2}{*}{$\downarrow$} 	& \textbf{11.792}	& 16.185	& 18.172 	& 34.635    & 10.503	& \textbf{10.444}	& 47.191	& 16.507     \\ 
              $L_2$ Norm						& 								& \textbf{1,678.2}	& 2,224.8	& 3,109.0 	& 4,749.2   & 2,368.5	& 2,342.5	& 6,661.3	& \textbf{2,319.8}    \\  \hline
              Absolute Diff						&\multirow{2}{*}{$\downarrow$} 	& \textbf{0.1019}	& 0.1441	& 0.1512 	& 1.2020    & \textbf{0.1838}	& 0.1936	& 0.9062	& 0.2836     \\
              Squared Diff						& 								& \textbf{2.9974}	& 5.3891	& 8.6444 	& 102.26	& \textbf{8.7122}	& 9.0332	& 100.89	& 9.3032     \\ \hline
              RMSE\textsubscript{lin}			&\multirow{3}{*}{$\downarrow$} 	& \textbf{18.677}	& 25.213	& 33.526 	& 49.973    & 24.865	& 24.699	& 72.084	& \textbf{24.521}     \\
              RMSE\textsubscript{log}			& 								& \textbf{0.1744}	& 0.2752	& 1.0864 	& 0.8330    & \textbf{0.2932}	& 0.2970	& 1.2240	& 0.3390     \\
              RMSE\textsubscript{scale-inv}		& 								& \textbf{0.1345}	& 0.2018	& 1.0774 	& 0.7009    & 0.2687	& \textbf{0.2642}	& 1.1759	& 0.2867     \\ \hline
              $\delta < 1.25$					&\multirow{3}{*}{$ \uparrow$} 	& \textbf{0.8529}	& 0.6854	& 0.7802 	& 0.4848    & \textbf{0.7393}	& 0.7230	& 0.4149	& 0.6395     \\
              $\delta < 1.25^2$					& 								& \textbf{0.9642}	& 0.8728	& 0.8978 	& 0.6332    & \textbf{0.9224}	& 0.9064	& 0.5298	& 0.7943     \\
              $\delta < 1.25^3$					& 								& \textbf{0.9915}	& 0.9651	& 0.9638 	& 0.7225    & \textbf{0.9609}	& 0.9557	& 0.6360	& 0.9311     \\ \hline 
			  $\textbf{Face Verification}$		&$\uparrow$						& \textbf{0.7247}	& 0.6570	& 0.5315 	& 0.6442    & \textbf{0.6251}	& 0.6043	& 0.5422	& 0.5966\ 
          \end{tabular}
      \end{center}
      \caption{Accuracy comparison for the \textit{pixel-wise} metrics and the \textit{face verification} task, as a function of different generative approaches. \textit{cGAN} and \textit{AE} refer to the generative adversarial network and the autoencoder proposed in this paper, respectively. Results are reported for both the \textit{Pandora} and \textit{Biwi} datasets and for the main competitors \cite{isola2017image,cui2018improving}.
      Starting from the top, $L_1$ and $L_2$ distances are reported, then the absolute and the squared differences, the root mean squared error, and the percentage of pixels under a certain error threshold. Further details about metrics are reported in \cite{eigen2014depth}. Finally, the accuracy on the face verification task, detailed in Section~\ref{sec:face-verification}, is reported. The arrows next to the metrics represent the positive changing direction: a better generation performance corresponds to a metric variation in the arrow direction.}
      \label{tab:metrics}
\end{table*}

\subsection{Datasets}
A description of the exploited datasets, namely \textit{Pandora} and \textit{Biwi Kinect Head Pose}, is provided. The explanation of how the \textit{Pandora} dataset was split to take different head poses, occlusions, and garments into account is presented in the following as well.

\subsubsection{Pandora Dataset}\label{sec:pandora}
The Pandora dataset was introduced in~\cite{borghi17cvpr} for the head pose estimation task in depth images. It consists of more than $250$k paired face images, both in the RGB and the depth domain.
Depth maps are acquired with the \textit{Microsoft Kinect One} device (also known as \textit{Microsoft Kinect for Windows v2}), a \textit{Time-of-Flight} sensor that assures great quality and high resolution for both the RGB ($1920 \times 1080$ pixels) and the depth ($512 \times 424$ pixels) data.
Even though the dataset was not created for the depth generation task, it can be successfully employed for that purpose as well, as it contains paired RGB-depth images.\\
Furthermore, it includes some challenging features, such as the presence of garments, numerous face occlusions created by objects (\eg bottles, smartphones, and tablets) and arms, and extreme head poses (\textit{roll}: $\pm \ang{70}$, \textit{pitch}: $\pm \ang{100}$, \textit{yaw}: $\pm \ang{125}$).
As reported in the original paper, we use subjects number 10, 14, 16, and 20 as a testing subset. \\
Each subject presents $5$ different sequences $S_i$ of frames. We split the sequences into two sets. The first one, referred as $\{S_1, S_2, S_3\}$, contains actions performed with constrained movements (yaw, pitch, and roll vary one at a time), for both the head and the shoulders. The second set, referred as $\{S_4, S_5\}$, consists of both complex and simple movements, as well as occlusions and challenging camouflage. Experiments are performed on both the subsets in order to investigate the effects of the mentioned differences. 
Moreover, we additionally split the dataset taking head pose angles into account. We create two mutually-exclusive head pose-based subsets, defined as
\begin{subequations}
	\label{eq:angles}
	\begin{equation}
		A_1 = \big\{ s_{\rho \theta \sigma} \,\vert\, \forall \gamma \in \{\rho, \theta, \sigma\} : \ang{-10} \leq \gamma \leq \ang{10} \big\} \label{eq:angles:a1}
        \end{equation}
        \begin{equation}
		A_2 = \big\{ s_{\rho \theta \sigma} \,\vert\, \exists \gamma \in \{\rho, \theta, \sigma\} : \gamma < \ang{-10} \lor \gamma > \ang{10} \big\} \label{eq:angles:a2}
        \end{equation}

\end{subequations}
where $\rho, \theta$, and $\sigma$ are the yaw, the pitch, and the roll angle, respectively, for each sample $s_{\rho \theta \sigma}$.
In practice, $A_1$ consists of frontal face images, while non-frontal face images are included in $A_2$.

When using the dataset for the face verification task, the problem of the high number of possible dataset image pairs arises. To overcome the issue, we created two fixed set of image pairs, a validation and a test set, in order to allow repeatable and comparable experiments.

We extract face images from dataset frames using the automatic face cropping technique presented in Section~\ref{sec:face-crop} then we resize them to the size of $96 \times 96$ pixels.
We exclude from the dataset a very small subset of extreme head poses and occlusions, as well as frames in which the automatic cropping algorithm does not work properly, to avoid training instability.

{
\begin{table*}
    \begin{center}
    	\begin{tabular}[th]{c|cc|cc|cc}
				& \multicolumn{2}{c|}{$\boldsymbol{\{S_i\}_{i = 1,2,3}}$} 	& \multicolumn{2}{c|}{$\boldsymbol{\{S_i\}_{i = 4,5}}$} 	& \multicolumn{2}{c}{$\boldsymbol{\{S_i\}_{i = 1,2,3,4,5}}$}		\\ 
            						& \textit{~original~}		& \textit{generated} 			& \textit{~original~}		& \textit{generated} 			& \textit{~original~}		& \textit{generated} 			\\  \hline
            \rule{0pt}{2ex}$\boldsymbol{A_1}$	& 0.8184	& 0.8614		& 0.7685	& 0.7155		& 0.7917	& 0.7950		\\ 
            $\boldsymbol{A_2}$					& 0.7928	& 0.7499		& 0.7216	& 0.6586		& 0.7576	& 0.7007		\\ 
        	$\boldsymbol{\{A_1, A_2\}}$			& 0.8034	& 0.7851		& 0.7271	& 0.6696		& 0.7664	& 0.7247		
		\end{tabular}
	\end{center}
    \caption{Confusion matrix for the Face Verification task on the \textit{Pandora} dataset, as a function of different angles and sequences subsets on original and generated depth images. Subsets description is reported in Section \ref{sec:pandora}. Tested generated images are estimated by the proposed network trained according to the adversarial approach described in Section \ref{sec:proposedmethod} for 30 epochs.
    }
    \label{tab:subsets}
\end{table*}
}

\subsubsection{Biwi Kinect Head Pose Database} \label{sec:biwi dataset}
The \textit{Biwi Kinect Head Pose Database} was introduced by Fanelli \etal~\cite{fanelli_IJCV} in 2013. Differently from Pandora, it is acquired with the first version of the \textit{Microsoft Kinect}, a \textit{structured-light} infrared sensor.
With respect to Tof sensors, this Microsoft Kinect version provides lower quality depth maps~\cite{sarbolandi2015kinect}, in which it is common to find holes (missing depth values).\\
The dataset consists of about 15k frames, split in $24$ sequences of $20$ different subjects (four subjects are recorded twice). Both RGB and depth images have the same spatial resolution of $640 \times 480$ pixels. 
The head pose angles span about $\pm \ang{50}$ for \textit{roll}, $\pm \ang{60}$ for \textit{pitch}, and $\pm \ang{75}$ for \textit{yaw}.
We adopt the same procedure used for Pandora to crop the faces and obtain $96 \times 96$ pixel images.

\subsection{Pixel-wise metrics} \label{sec:pixel wise metrics}
The overall quality of the generated facial depth maps is evaluated with the pixel-wise metrics proposed in \cite{eigen2014depth}. In particular, the generation capability of the generator network, trained both as an autoencoder and with the adversarial policy reported in Section~\ref{sec:adversarial-training}, is compared with the recent \textit{pix2pix} architecture~\cite{isola2017image} and the algorithm proposed in \cite{cui2018improving}. We test the models on both the \textit{Pandora} and the \textit{Biwi} dataset. \\
As reported in Table \ref{tab:metrics}, our generator network, trained as a cGAN, performs better than the autoencoder and the literature competitors.
As highlighted in right part of the above-mentioned table, the limited size and variability of the Biwi dataset have a negative impact on the generative and the generalization capability of the tested architectures.
Nevertheless, the $\delta$-metrics, corresponding to the percentage of pixels under a certain error threshold, confirm that the proposed model achieves the best spatial accuracy with a clear margin on both datasets.

\subsection{Face Verification test} \label{sec:face-verification}
Pixel-wise metrics allow for a mathematical evaluation of the generative performance of deep convolutional networks. Yet, they might not fully convey whether the original domain features are accurately preserved through the generative process. Even when a human observer perceives no difference between \enquote{real} and \enquote{fake} images, the information content might still be represented in a slightly different fashion in terms of texture, colors, geometries, light intensity, and fine details.

In order to deeply investigate the quality of the generated images, the following Face Verification test, \ie~determining whether two given face images belong to the same subject, is employed. 
We exploit a deep convolutional Siamese network trained on original depth images, without adopting any kind of fine-tuning on the generated depth maps. The model compares two depth images and predicts their similarity as a value in the $[0, 1]$ range. Two input faces are considered as belonging to the same person if their similarity score is higher than $0.5$. \\
The model architecture is depicted in Figure \ref{fig:architecture_siam}. It is composed by $5$ convolutional layers with an increasing number of feature maps, an average pooling layer, and $3$ fully connected layers.

\begin{figure*}[t]
  \centering
  \includegraphics[width=0.9\linewidth]{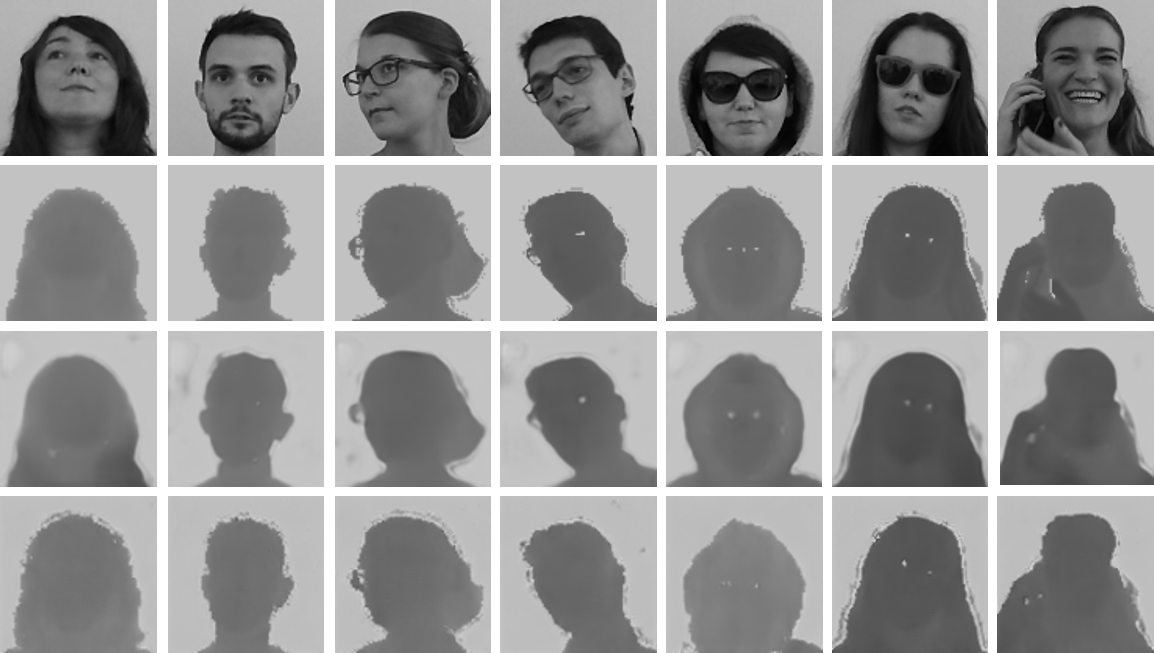} 
  \caption{Sample outputs of the proposed method (last row). In the first and second rows, original gray-level and depth maps are reported, respectively. In the third row, depth maps generated by the autoencoder are depicted.}
  \label{fig:results}
\end{figure*}

Through this test, which needs both high and low level features in order to work properly, we can estimate how well the generated faces preserve the individual visual features of the subjects. 
As reported in the last line of Table \ref{tab:metrics}, the proposed cGAN model allows for the highest test accuracy on the face verification task. Actually, the proposed architecture significantly outperforms the other tested architectures. \\
In light of the reported results, we believe that our model is able to estimate both high level and fine details of facial depth maps and hence to obtain realistic and discriminative synthetic images.\\
It is worth to notice that our architecture overcomes the model proposed in \cite{cui2018improving}, in which the deep model is specifically trained to preserve subject identities by minimizing a dedicated face recognition loss.

Table \ref{tab:subsets} presents how the Face Verification accuracy varies as a function of different head poses and image complexity (\textit{i.e.} obstructions, garments, unconstrained movements) on the \textit{Pandora} dataset, as reported in Section \ref{sec:pandora}. Furthermore, we report the verification accuracy obtained by the same model on the original depth maps.\\
As expected, the Siamese network performs better on frontal faces (angle subset $A_1$) and constrained movements (sequence subset $\{S_1, S_2, S_3\}$). Nevertheless, even the most complex task ($A_2$, $\{S_4, S_5\}$) reaches a $72.47\%$ accuracy.
Surprisingly, in the case of frontal faces and constrained movements, the generated depth images lead to better accuracy than the original ones. We hypothesize that this behavior is related to the generative process that tends to remove partial occlusions and to produce highly discriminative facial features.

\section{Conclusion}
In this paper, we present an approach for the estimation of facial depth maps from intensity images.
In order to evaluate the quality of the generated images, we perform a face verification task employing a Siamese network pre-trained on the original depth maps.
By showing that the Siamese network accuracy does not degrade when tested on generated images, we demonstrate that the proposed framework produces high-quality depth maps, both in terms of visual appearance and discriminative information.
We also demonstrate that the proposed architecture outperforms the autoencoder and the literature competitors when trained with the adversarial policy. \\
Thanks to the flexibility of our approach, we plan to extend our model by introducing task-specific losses and to apply it to different scenarios.

{\small
\bibliographystyle{ieee}
\bibliography{egbib}

\begin{thebibliography}{10}\itemsep=-1pt

\bibitem{atapour2018real}
A.~Atapour-Abarghouei and T.~Breckon.
\newblock Real-time monocular depth estimation using synthetic data with domain
  adaptation via image style transfer.
\newblock In {\em Proceedings of the IEEE Conference on Computer Vision and
  Pattern Recognition}, volume~18, 2018.

\bibitem{baig2014im2depth}
M.~H. Baig, V.~Jagadeesh, R.~Piramuthu, A.~Bhardwaj, W.~Di, and N.~Sundaresan.
\newblock Im2depth: Scalable exemplar based depth transfer.
\newblock In {\em Applications of Computer Vision (WACV), 2014 IEEE Winter
  Conference on}, pages 145--152. IEEE, 2014.

\bibitem{borghi17cvpr}
G.~Borghi, M.~Venturelli, R.~Vezzani, and R.~Cucchiara.
\newblock Poseidon: Face-from-depth for driver pose estimation.
\newblock In {\em IEEE International Conference on Computer Vision and Pattern
  Recognition}. IEEE, 2017.

\bibitem{cavestany2015improved}
P.~Cavestany, A.~L. Rodr{\'\i}guez, H.~Mart{\'\i}nez-Barber{\'a}, and T.~P.
  Breckon.
\newblock Improved 3d sparse maps for high-performance sfm with low-cost
  omnidirectional robots.
\newblock In {\em Image Processing (ICIP), 2015 IEEE International Conference
  on}, pages 4927--4931. IEEE, 2015.

\bibitem{cui2018improving}
J.~Cui, H.~Zhang, H.~Han, S.~Shan, and X.~Chen.
\newblock Improving 2d face recognition via discriminative face depth
  estimation.
\newblock {\em Proc. ICB}, pages 1--8, 2018.

\bibitem{ding2017fusing}
L.~Ding and G.~Sharma.
\newblock Fusing structure from motion and lidar for dense accurate depth map
  estimation.
\newblock In {\em Acoustics, Speech and Signal Processing (ICASSP), 2017 IEEE
  International Conference on}, pages 1283--1287. IEEE, 2017.

\bibitem{eigen2015predicting}
D.~Eigen and R.~Fergus.
\newblock Predicting depth, surface normals and semantic labels with a common
  multi-scale convolutional architecture.
\newblock In {\em Proceedings of the IEEE International Conference on Computer
  Vision}, pages 2650--2658, 2015.

\bibitem{eigen2014depth}
D.~Eigen, C.~Puhrsch, and R.~Fergus.
\newblock Depth map prediction from a single image using a multi-scale deep
  network.
\newblock In {\em Advances in neural information processing systems}, pages
  2366--2374, 2014.

\bibitem{fanelli_IJCV}
G.~Fanelli, M.~Dantone, J.~Gall, A.~Fossati, and L.~Van~Gool.
\newblock Random forests for real time 3d face analysis.
\newblock {\em Int. J. Comput. Vision}, 101(3):437--458, February 2013.

\bibitem{garg2016unsupervised}
R.~Garg, V.~K. BG, G.~Carneiro, and I.~Reid.
\newblock Unsupervised cnn for single view depth estimation: Geometry to the
  rescue.
\newblock In {\em European Conference on Computer Vision}, pages 740--756.
  Springer, 2016.

\bibitem{godard2017unsupervised}
C.~Godard, O.~Mac~Aodha, and G.~J. Brostow.
\newblock Unsupervised monocular depth estimation with left-right consistency.
\newblock In {\em CVPR}, volume~2, page~7, 2017.

\bibitem{goodfellow2014generative}
I.~Goodfellow, J.~Pouget-Abadie, M.~Mirza, B.~Xu, D.~Warde-Farley, S.~Ozair,
  A.~Courville, and Y.~Bengio.
\newblock Generative adversarial nets.
\newblock In {\em Advances in neural information processing systems}, pages
  2672--2680, 2014.

\bibitem{iizuka2016let}
S.~Iizuka, E.~Simo-Serra, and H.~Ishikawa.
\newblock Let there be color!: joint end-to-end learning of global and local
  image priors for automatic image colorization with simultaneous
  classification.
\newblock {\em ACM Transactions on Graphics (TOG)}, 35(4):110, 2016.

\bibitem{isola2017image}
P.~Isola, J.~Zhu, T.~Zhou, and A.~A. Efros.
\newblock Image-to-image translation with conditional adversarial networks.
\newblock In {\em {IEEE} Conference on Computer Vision and Pattern Recognition,
  {CVPR}}, 2017.

\bibitem{kingma2014adam}
D.~P. Kingma and J.~Ba.
\newblock Adam: A method for stochastic optimization.
\newblock {\em arXiv preprint arXiv:1412.6980}, 2014.

\bibitem{konda2013unsupervised}
K.~Konda and R.~Memisevic.
\newblock Unsupervised learning of depth and motion.
\newblock {\em arXiv preprint arXiv:1312.3429}, 2013.

\bibitem{kong2016effective}
D.~Kong, Y.~Yang, Y.-X. Liu, M.~Li, and H.~Jia.
\newblock Effective 3d face depth estimation from a single 2d face image.
\newblock In {\em Communications and Information Technologies (ISCIT), 2016
  16th International Symposium on}, pages 221--230. IEEE, 2016.

\bibitem{ladicky2014pulling}
L.~Ladicky, J.~Shi, and M.~Pollefeys.
\newblock Pulling things out of perspective.
\newblock In {\em Proceedings of the IEEE Conference on Computer Vision and
  Pattern Recognition}, pages 89--96, 2014.

\bibitem{laina2016deeper}
I.~Laina, C.~Rupprecht, V.~Belagiannis, F.~Tombari, and N.~Navab.
\newblock Deeper depth prediction with fully convolutional residual networks.
\newblock In {\em 3D Vision (3DV), 2016 Fourth International Conference on},
  pages 239--248. IEEE, 2016.

\bibitem{larsson2016learning}
G.~Larsson, M.~Maire, and G.~Shakhnarovich.
\newblock Learning representations for automatic colorization.
\newblock In {\em European Conference on Computer Vision}, pages 577--593.
  Springer, 2016.

\bibitem{li2015depth}
B.~Li, C.~Shen, Y.~Dai, A.~van~den Hengel, and M.~He.
\newblock Depth and surface normal estimation from monocular images using
  regression on deep features and hierarchical crfs.
\newblock In {\em Proceedings of the IEEE Conference on Computer Vision and
  Pattern Recognition}, pages 1119--1127, 2015.

\bibitem{li2017two}
J.~Li, R.~Klein, and A.~Yao.
\newblock A two-streamed network for estimating fine-scaled depth maps from
  single rgb images.
\newblock In {\em Proceedings of the IEEE Conference on Computer Vision and
  Pattern Recognition}, pages 3372--3380, 2017.

\bibitem{liu2016learning}
F.~Liu, C.~Shen, G.~Lin, and I.~Reid.
\newblock Learning depth from single monocular images using deep convolutional
  neural fields.
\newblock {\em IEEE transactions on pattern analysis and machine intelligence},
  38(10):2024--2039, 2016.

\bibitem{liu2016coupled}
M.-Y. Liu and O.~Tuzel.
\newblock Coupled generative adversarial networks.
\newblock In {\em Advances in neural information processing systems}, pages
  469--477, 2016.

\bibitem{long2015fully}
J.~Long, E.~Shelhamer, and T.~Darrell.
\newblock Fully convolutional networks for semantic segmentation.
\newblock In {\em Proceedings of the IEEE Conference on Computer Vision and
  Pattern Recognition}, pages 3431--3440, 2015.

\bibitem{maas2013rectifier}
A.~L. Maas, A.~Y. Hannun, and A.~Y. Ng.
\newblock Rectifier nonlinearities improve neural network acoustic models.
\newblock In {\em ICML Workshop on Deep Learning for Audio, Speech, and
  Language Processing}, 2013.

\bibitem{masci2011stacked}
J.~Masci, U.~Meier, D.~Cire{\c{s}}an, and J.~Schmidhuber.
\newblock Stacked convolutional auto-encoders for hierarchical feature
  extraction.
\newblock In {\em International Conference on Artificial Neural Networks},
  pages 52--59. Springer, 2011.

\bibitem{mirza2014conditional}
M.~Mirza and S.~Osindero.
\newblock Conditional generative adversarial nets.
\newblock {\em arXiv preprint arXiv:1411.1784}, 2014.

\bibitem{reiter2006estimation}
M.~Reiter, R.~Donner, G.~Langs, and H.~Bischof.
\newblock {\em Estimation of face depth maps from color textures using
  canonical correlation analysis}.
\newblock na, 2006.

\bibitem{sarbolandi2015kinect}
H.~Sarbolandi, D.~Lefloch, and A.~Kolb.
\newblock Kinect range sensing: Structured-light versus time-of-flight kinect.
\newblock {\em Computer Vision and Image Understanding}, 2015.

\bibitem{saxena2006learning}
A.~Saxena, S.~H. Chung, and A.~Y. Ng.
\newblock Learning depth from single monocular images.
\newblock In {\em Advances in neural information processing systems}, pages
  1161--1168, 2006.

\bibitem{saxena2009make3d}
A.~Saxena, M.~Sun, and A.~Y. Ng.
\newblock Make3d: Learning 3d scene structure from a single still image.
\newblock {\em IEEE transactions on pattern analysis and machine intelligence},
  31(5):824--840, 2009.

\bibitem{sun2011depth}
Z.-L. Sun and K.-M. Lam.
\newblock Depth estimation of face images based on the constrained ica model.
\newblock {\em IEEE Transactions on Information Forensics and Security},
  6(2):360--370, 2011.

\bibitem{sun2013depth}
Z.-L. Sun, K.-M. Lam, and Q.-W. Gao.
\newblock Depth estimation of face images using the nonlinear least-squares
  model.
\newblock {\em IEEE transactions on image processing}, 22(1):17--30, 2013.

\bibitem{tao2015depth}
M.~W. Tao, P.~P. Srinivasan, J.~Malik, S.~Rusinkiewicz, and R.~Ramamoorthi.
\newblock Depth from shading, defocus, and correspondence using light-field
  angular coherence.
\newblock In {\em Proceedings of the IEEE Conference on Computer Vision and
  Pattern Recognition}, pages 1940--1948, 2015.

\bibitem{wang2016generative}
X.~Wang and A.~Gupta.
\newblock Generative image modeling using style and structure adversarial
  networks.
\newblock In {\em European Conference on Computer Vision}, pages 318--335.
  Springer, 2016.

\bibitem{woodham1980photometric}
R.~J. Woodham.
\newblock Photometric method for determining surface orientation from multiple
  images.
\newblock {\em Optical engineering}, 19(1):191139, 1980.

\bibitem{xie2015holistically}
S.~Xie and Z.~Tu.
\newblock Holistically-nested edge detection.
\newblock In {\em Proceedings of the IEEE international conference on computer
  vision}, pages 1395--1403, 2015.

\bibitem{xu2017multi}
D.~Xu, E.~Ricci, W.~Ouyang, X.~Wang, and N.~Sebe.
\newblock Multi-scale continuous crfs as sequential deep networks for monocular
  depth estimation.
\newblock In {\em Proceedings of CVPR}, 2017.

\bibitem{yamaguchi2012continuous}
K.~Yamaguchi, T.~Hazan, D.~McAllester, and R.~Urtasun.
\newblock Continuous markov random fields for robust stereo estimation.
\newblock In {\em European Conference on Computer Vision}, pages 45--58.
  Springer, 2012.

\bibitem{zhao2016energy}
J.~Zhao, M.~Mathieu, and Y.~LeCun.
\newblock Energy-based generative adversarial network.
\newblock {\em arXiv preprint arXiv:1609.03126}, 2016.

\bibitem{zheng2008robust}
Y.~Zheng and Z.~Wang.
\newblock Robust depth estimation for efficient 3d face reconstruction.
\newblock In {\em Image Processing, 2008. ICIP 2008. 15th IEEE International
  Conference on}, pages 1516--1519. IEEE, 2008.

\bibitem{zhou2017unsupervised}
T.~Zhou, M.~Brown, N.~Snavely, and D.~G. Lowe.
\newblock Unsupervised learning of depth and ego-motion from video.
\newblock In {\em CVPR}, volume~2, page~7, 2017.

\end{thebibliography}
}

\end{document}